\documentclass[11pt]{article}

\usepackage[preprint]{acl}

\usepackage{times}
\usepackage{latexsym}
\usepackage{amsmath}
\usepackage{amsfonts}
\usepackage{listings,xcolor}
\lstdefinestyle{prompt}{
  basicstyle=\ttfamily\scriptsize,
  breaklines=true, breakatwhitespace=true, columns=fullflexible,
  keepspaces=true, tabsize=2,
  frame=single, xleftmargin=-1em, xrightmargin=2em,
  postbreak=\mbox{\textcolor{gray}{\(\hookrightarrow\)}\space}
}

\usepackage{booktabs}
\usepackage[table]{xcolor} 
\usepackage{multirow}
\usepackage{graphicx} 

\usepackage[T1]{fontenc}

\usepackage[utf8]{inputenc}

\usepackage{algorithm} 
\usepackage{algpseudocode}
\usepackage{xcolor}
\definecolor{cinit}{HTML}{7F7F7F}     
\definecolor{cdesign}{HTML}{1F77B4}   
\definecolor{ccode}{HTML}{2CA02C}     
\definecolor{cexec}{HTML}{17BECF}     
\definecolor{creflect}{HTML}{FF7F0E}  
\definecolor{carchive}{HTML}{9467BD}  
\colorlet{cdesign}{cdesign!50}    
\colorlet{ccode}{ccode!50}
\colorlet{cexec}{cexec!50}
\colorlet{creflect}{creflect!50}
\colorlet{carchive}{carchive!55}
\newcommand{\steptag}[2]{\textcolor{#1}{[#2]}}
\usepackage{amssymb}

\usepackage{microtype}

\usepackage{inconsolata}
\usepackage{hyperref}

\usepackage{graphicx}

\title{MedDCR: Learning to Design Agentic Workflows for Medical Coding}

\author{Jiyang Zheng\textsuperscript{1,2}, Islam Nassar\textsuperscript{1}, Thanh Vu\textsuperscript{1},  Xu Zhong\textsuperscript{1}, \\ \textbf{ Yang Lin\textsuperscript{1}, Tongliang Liu\textsuperscript{2}, Long Duong\textsuperscript{1}, Yuan-Fang Li\textsuperscript{1}} \\
\textsuperscript{1}Oracle Health and AI \\ \textsuperscript{2}Sydney AI Center, The University of Sydney\\
\texttt{\{jiyang.zheng, islam.nassar, thanh.v.vu, peter.zhong,}\\ 
\texttt{yang.y.lin, long.duong, yuanfang.li\}@oracle.com}\\
\texttt{\{tongliang.liu\}@sydney.edu.au}\\
}

\begin{document}
\maketitle
\begin{abstract}
Medical coding converts free-text clinical notes into standardized diagnostic and procedural codes, which are essential for billing, hospital operations, and medical research. Unlike ordinary text classification, it requires multi-step reasoning: extracting diagnostic concepts, applying guideline constraints, mapping to hierarchical codebooks, and ensuring cross-document consistency. Recent advances leverage agentic LLMs, but most rely on rigid, manually crafted workflows that fail to capture the nuance and variability of real-world documentation, leaving open the question of how to systematically learn effective workflows. We present \emph{\textbf{MedDCR}}, a closed-loop framework that treats workflow design as a learning problem. A \underline{D}esigner proposes workflows, a \underline{C}oder executes them, and a \underline{R}eflector evaluates predictions and provides constructive feedback, while a memory archive preserves prior designs for reuse and iterative refinement. On benchmark datasets, MedDCR outperforms state-of-the-art baselines and produces interpretable, adaptable workflows that better reflect real coding practice, improving both the reliability and trustworthiness of automated systems.
\end{abstract}

\section{Introduction}
\label{sec:introduction}
Medical coding is the process of translating unstructured clinical notes into standardized diagnostic and procedural codes, most commonly following the World Health Organization’s International Classification of Diseases (ICD) standard~\cite{dong2022automated}. These codes underpin critical functions in healthcare, including billing and reimbursement, hospital resource planning, and epidemiological research~\cite{campbell2020computer}. Unlike simple classification, manual coding requires coders to engage in a multi-step workflow: identifying relevant mentions in free text, consulting multiple resources such as the alphabetic index and tabular index, applying coding guidelines, and cross-checking for consistency across diagnoses and procedures. This structured but intricate process makes medical coding highly labour-intensive and error-prone, contributing to global coding backlogs and clinical risks when errors occur~\cite{alonso2020problems, douglas2025less, gan-etal-2025-aligning}.

To reduce the burden of manual coding, recent research has leveraged large language models (LLMs) as the foundation for automated systems~\cite{boyle2023automated, yang2023surpassing, falis2024can, baksi2025medcoder}. Beyond their core ability to map free-text clinical notes to candidate codes, LLMs possess reasoning and tool-use capabilities that are particularly well-suited to the structured, rule-based nature of the coding process~\cite{kwan2024large, mustafa2025evaluating}. 

Building on this, emerging work has proposed agentic coding workflows~\cite{li2024exploring, motzfeldt2025code}, where multiple interacting agents cooperate to mirror the steps taken by human coders: extracting relevant terms, consulting indexes and guidelines, and verifying consistency across diagnoses and procedures. This paradigm has demonstrated competitive performance and enhanced interpretability compared to treating coding as a flat multi-label classification problem.

\begin{figure*}[t]
  \includegraphics[width=\textwidth, page=1]{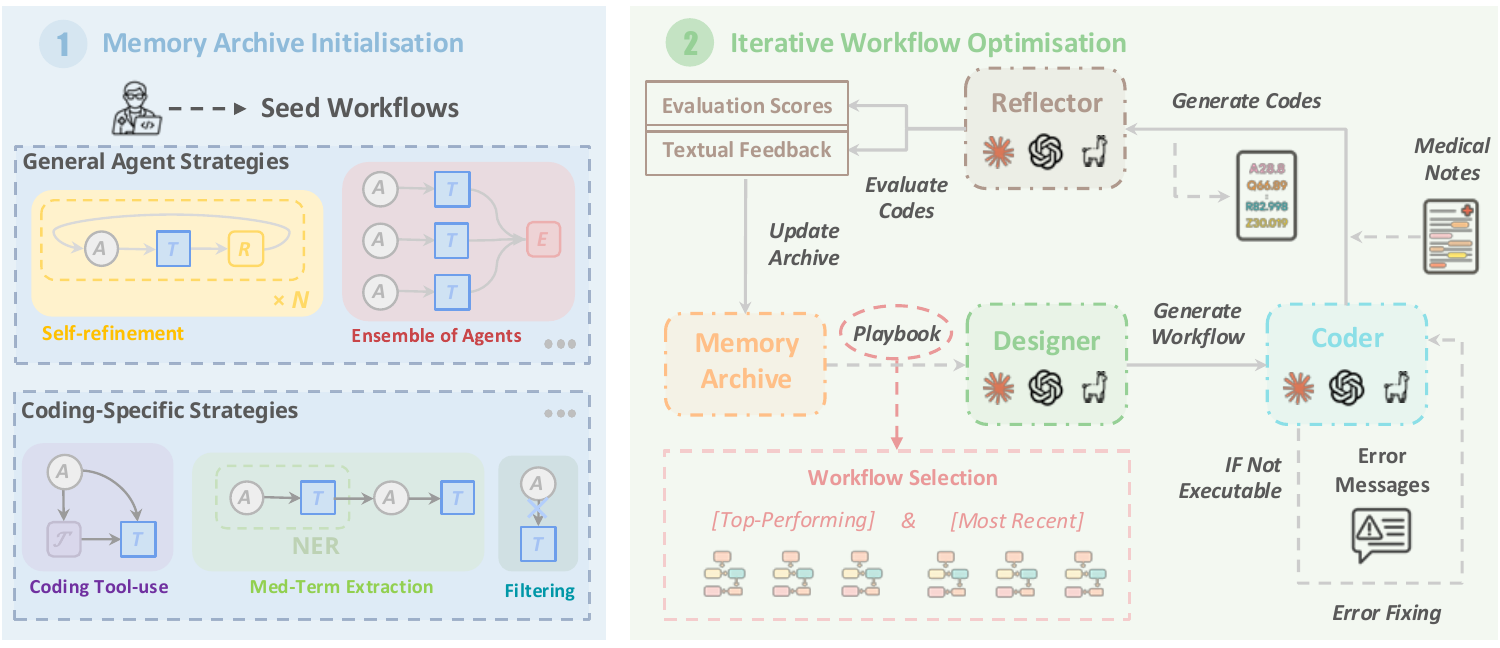}
  \caption{Overview of the \textbf{MedDCR} framework. (1) The \textit{memory archive} is initialised with general reasoning strategies (e.g., self-refinement, multi-agent ensembles, chain-of-thought prompting) and coding-specific strategies (e.g., medical term extraction, weak code filtering, ICD tool use), together with other optional seed workflows. (2) In each optimisation loop, the \textit{Designer} proposes new workflows, the \textit{Coder} compiles and executes them (with self-fixing if needed), and the \textit{Reflector} provides both evaluation scores and textual feedback. The memory archive stores all past workflows, enabling reuse, progressive refinement, and workflow selection from top-performing and recent designs. This closed-loop process discovers effective coding workflows under guideline constraints.}
  \label{fig:MedDCR_framework}
\end{figure*}

Despite these advances, most existing agentic frameworks for medical coding remain manually crafted, relying on human experts to specify the modules and execution steps within a workflow~\cite{motzfeldt2025code}. Yet this \textit{design problem} is particularly challenging, as correct code assignment with LLM-based systems often hinges on two factors: (i) the quality of module definitions (e.g., tool calls, inference strategies), and (ii) the effectiveness of interactions across modules (i.e., how tools and strategies are combined and ordered). Manually fixing these design choices risks overlooking more effective coordination patterns, thereby limiting the potential of agentic approaches~\cite{li2024more, zhang2025aflow}. For instance, automated search may discover non-obvious yet beneficial strategies, such as applying a \textit{contrastive screening} step to prune near-duplicate ICD codes based on description similarity, whereas manual designs may fail to recognize such refinements. This motivates the need for automated workflow learning, which can flexibly search for and refine workflow designs rather than constraining systems to static, expert-defined pipelines.

To address this challenge, we propose MedDCR (As demonstrated in Figure~\ref{fig:MedDCR_framework}), an automated framework that optimises workflows for medical coding. Instead of relying on a single fixed pipeline, MedDCR treats workflow design as a learning problem~\cite{hu2025automated,zhang2025multiagent, zhang2025aflow, zhou2025multi}, where workflows are proposed, executed, and evaluated in an iterative loop. 

Within this loop, a \textbf{\underline{D}}esigner agent generates workflow plans by inventing or combining coding tools and strategies. A \textbf{\underline{C}}oder agent then translates these plans into executable pipelines in the form of concrete programs, which carry out operations such as tool calling, validation, and reconciliation to predict medical codes. Finally, a \textbf{\underline{R}}eflector agent evaluates the predictions, providing both performance scores and textual feedback on the effectiveness of the workflow design. The Designer uses this feedback to refine its proposals, enabling workflows to evolve and improve over time~\cite{wang2025maestro}. Beyond this loop, MedDCR maintains a memory archive of prior designs, enabling workflows to be reused, progressively refined, or augmented with expert-crafted pipelines~\cite{ji2024unified, li2024exploring, motzfeldt2025code}. 


Our experiments demonstrate that the workflows discovered by MedDCR significantly outperform both state-of-the-art, hand-designed baselines and pretrained language model approaches. Specifically, MedDCR achieves a 6.2\% improvement in F1 score on the MDACE~\cite{cheng-etal-2023-mdace} dataset and a 7.4\% gain on ACI-BENCH~\cite{aci-bench}, compared to the second-best performing baselines. 

Building upon these insights and findings, we present our main contributions as follows:
\begin{itemize}
    \item We propose MedDCR, a novel framework for medical coding that treats workflow design as a learning problem through iterative design–coding/execute–reflect loops.

    \item We develop a meta-agent architecture with Designer, Coder, and Reflector agents, supported by a memory archive that enables both search from scratch and plug-and-play optimisation of expert medical coding workflows.

    \item MedDCR demonstrates state-of-the-art performance on multiple benchmarks, along with improved interpretability and risk analysis.
\end{itemize}

\section{Related Works}
\label{sec:related}
\subsection{Automated Medical Coding}
Automated medical coding seeks to accelerate the mapping of free-text clinical notes to standardized medical codes using computer-assisted tools. Early systems were rule-based~\cite{farkas2008automatic, kavuluru2015empirical}, but the availability of large EHR datasets such as MIMIC-III and MIMIC-IV~\cite{johnson2016mimic, johnson2020mimic} established deep learning models as the dominant approach, typically framing coding as an extreme multi-label classification task. Encoder–decoder architectures with label-wise attention~\cite{mullenbach2018explainable, li2020icd, vu2021label} were extended with textual descriptions, synonyms, or co-occurrence signals to better align notes and codes~\cite{cao2020clinical, dong2021explainable, yuan2022code}. More recently, pre-trained language models fine-tuned for ICD prediction (PLM-ICD) achieved state-of-the-art performance~\cite{huang2022plm, edin2024unsupervised, douglas2025less}. However, these models struggle with the extremely large ICD label space, rare codes, and long clinical notes.

To address these challenges, researchers have explored LLMs for generative coding. Zero and few-shot prompting~\cite{yang2023multi, boyle2023automated, geroself} showed flexibility but underperformed PLM-based classifiers. This has motivated agentic approaches, where LLMs interact with external tools, indexes, and validation routines in multi-agent workflows that mimic human coding processes~\cite{li2024exploring, kwan2024large, motzfeldt2025code}. While promising, these workflows remain manually designed and fixed, making them potentially suboptimal.

Our work addresses this gap by introducing MedDCR, which treats workflow design as a learning problem and automatically searches for effective agentic workflows for medical coding, enhancing both predictive performance and interpretability.

\subsection{Agentic Workflow Design}
Agentic workflows frame problem solving as a coordinated process across one or more LLM-based agents, each assigned specific roles or equipped with external tools. Recent advances enrich these systems with prompting strategies~\cite{chen2023unleashing}, chain-of-thought planning~\cite{wei2022chain, zheng2025chainoffocus}, reflection and refinement~\cite{madaan2023self, shinn2023reflexion, dhuliawala2024chain}, tool use~\cite{nakano2021webgpt, schick2023toolformer, zheng2025aligning}, and role assignment for multi-agent cooperation~\cite{shanahan2023role, li2023camel, wu2024autogen}. Multi-agent topologies vary from parallel setups for exploration~\cite{wangself} to serial pipelines with reflective refinement~\cite{madaan2023self}, and even contrastive-style structures that improve reliability~\cite{du2023improving, zheng2024enhancing}.

Building on these foundations, the community has begun exploring automated design of agentic systems. Most works focus on automating prompt optimization~\cite{yang2023large, fernando2024promptbreeder, khattab2024dspy}, role definition~\cite{li2023camel, wu2024autogen}, or specific topology search~\cite{zhou2025multi, wang2025maestro}. Others attempt to expand the search space to workflows~\cite{hu2025automated, zhang2025multiagent, zhang2025aflow}, where both the definition of workflow components and their topological organisation are jointly optimised, but in practice, many components remain fixed, making discovered agents less flexible.

Our work advances this line by introducing MedDCR, which treats agentic workflow design as a learning problem. Unlike prior approaches, MedDCR employs a meta-agent architecture with a Designer, Coder, and Reflector, reinforced by a memory archive that supports reuse and refinement of workflows. Coupled with medical coding tools and guideline-driven strategies, MedDCR provides the first domain-specific framework for automated optimisation of medical coding workflows, offering stronger performance and greater interpretability.
\section{Method}
\label{sec:method}
\subsection{Problem Formulation}
\label{sec:prob_form}
We formalize automated workflow optimisation problem for medical coding as follows.  
Let $\mathcal{X}$ denote the space of clinical notes and $\mathcal{C}$ the set of admissible ICD codes. A dataset is 
\begin{equation}
\begin{aligned}
    D=\{(x_i, Y_i)\}_{i=1}^n,\quad x_i \in \mathcal{X},\; Y_i \subseteq \mathcal{C}.
\end{aligned}
\end{equation}
A \emph{workflow} $W$ induces a coder function
\begin{equation}
\begin{aligned}
    f_W: \mathcal{X} \to 2^{\mathcal{C}},
\end{aligned}
\end{equation}
that maps a note to a predicted set of codes.  

Workflows are constructed from a component library $\mathcal{L}$ consisting of:  
(i) a tool set $\mathcal{T}$ (e.g., ICD index retrieval, guideline validators),  
(ii) reasoning/strategy primitives $\Sigma$ (e.g., extraction, validation, reconciliation), and  
(iii) LLM modules $\mathcal{M}$ with parameter configurations $\Theta$.  

We represent a workflow as a directed acyclic graph
\begin{equation}
\begin{aligned}
    W = (G=(V,E), \phi),
\end{aligned}
\end{equation}
where each node $v \in V$ instantiates a component $c_v \in \mathcal{L}$ with parameters $\phi_v \in \Theta$, and edges $E$ define data/control flow. The \emph{search space} of feasible workflows is
\begin{equation}
    \begin{aligned}
        S = \left\{ W = (G, \phi) \;\middle|\;
        \begin{array}{l}
          G \text{ is a DAG over } \mathcal{L}, \\
          W \text{ is executable}
        \end{array}
        \right\}.
    \end{aligned}
\end{equation}
Medical coding is constrained by official guidelines. Let $\Gamma$ denote the guideline resource (ICD Alphabetic/Tabular Index, coding rules). A compliance oracle $V_\Gamma(W,x) \in [0,1]$ measures the fraction of guideline checks that pass for workflow $W$ on input $x$. We evaluate a workflow with a task metric $g(\cdot,\cdot)$ (e.g., micro/macro $F_1$) and define the objective on a validation set $D_{\text{val}}$:
\begin{equation}
    \begin{aligned}
    G(W; D_{\text{val}}, \Gamma) 
    &= \mathbb{E}_{(x, Y) \sim D_{\text{val}}} \Big[ g(f_W(x), Y)  \\
    &\quad - \lambda_{\text{viol}} \big(1 - V_{\Gamma}(W, x)\big) \Big] \\
    &\quad - \lambda_{\text{cost}} \, C(W), 
    \end{aligned}
\end{equation}
where $C(W)$ measures resource usage (e.g., tool calls, latency, tokens) and $\lambda_{\mathrm{viol}},\lambda_{\mathrm{cost}} \geq 0$ control the trade-offs between predictive performance, guideline compliance, and efficiency.  

The automated workflow optimisation problem for medical coding is then:
\begin{equation}
    \begin{aligned}
        W^\star \in \arg\max_{W \in \mathcal{S}} \; G(W; D_{\text{val}}, \Gamma).
    \end{aligned}
\end{equation}

In the remainder of this section, we detail how MedDCR performs the iterative search for $W^\star$ guided by the objective $G$ and the constraints $\Gamma$.

\begin{algorithm}[t]
\caption{DCR Optimisation}
\label{alg:meddcr_summary}
\begin{algorithmic}[1]
\State \textbf{Input:} Validation set $D_{\text{val}}$, module library $\mathcal{L}$, objective $G$, iterations $T$, generation size M.
\State \textbf{Output:} Optimized workflow $W^\star$.
\State \steptag{cdesign}{Initialization of Archive}
\State Initialize memory archive $\mathcal{H}_0$ with optional, pre-evaluated seed workflows $\mathcal{W}_{\text{seed}}$.
\For{$t=1$ to $T$}
    \State \steptag{ccode}{Workflow Design Phase}
    \State \parbox[t]{\dimexpr\linewidth-\algorithmicindent}{Select elite $\mathcal{E}_t$ and recent $\mathcal{R}_t$ workflow examples from archive $\mathcal{H}_{t-1}$.}
    \vskip 1mm
    \State \parbox[t]{\dimexpr\linewidth-\algorithmicindent}{Propose a new set of workflow plans $\{\pi_t^{(m)}\}_{m=1}^M$ guided by $\mathcal{E}_t$ and $\mathcal{R}_t$.}
    \Statex
    \State \steptag{cexec}{Code \& Reflect Phase}
    \State \parbox[t]{\dimexpr\linewidth-\algorithmicindent}{Compile each plan $\pi_t$ into a runnable program of the workflow $W_t$.} \vskip 1mm
    \For{each workflow $W_t$}
        \State \parbox[t]{\dimexpr\linewidth-\algorithmicindent}{Execute $W_t$ on $D_{\text{val}}$ to obtain score $s_t$\\ and reflection $r_t$.}
        \State \parbox[t]{\dimexpr\linewidth-\algorithmicindent}{Update archive:\\
        $\mathcal{H}_t \leftarrow \mathcal{H}_{t-1} \cup \{(\pi_t, W_t, s_t, r_t)\}$.}
    \EndFor
\EndFor
\State \steptag{carchive}{Selection of Optimal Workflow}
\State \parbox[t]{\dimexpr\linewidth-\algorithmicindent}{Obtain best workflow\\ $W^\star \leftarrow \arg\max_{(\pi,W,s,r) \in \mathcal{H}_T} s$.}
\State \textbf{Return} optimized workflow $W^\star$.
\end{algorithmic}
\end{algorithm}

\subsection{MedDCR Framework Overview}
MedDCR framework instantiates the workflow optimisation problem defined in Section~\ref{sec:prob_form} by organising the search process into an iterative loop of design, coding (execution), and reflection. The framework aims to automatically discover effective workflows for medical coding without relying on fixed, manually crafted pipelines. 

The loop begins with the initialisation of a memory archive $\mathcal{H}_0$, which contains a small collection of seed workflows (See Figure~\ref{fig:MedDCR_framework}). These seeds capture both general reasoning strategies, such as chain-of-thought~\cite{wei2022chain, zhang2023automatic}, self-refinement~\cite{madaan2023self}, and multi-agent debate~\cite{du2023improving} and domain-specific medical coding strategies, including medical entity extractions~\cite{douglas2025less}, tabular index similarity checks, and reconciliation heuristics~\cite{gan-etal-2025-aligning}, among others. The archive also stores a list of coding tools, such as alphabetic index search, parent--child code lookup, and code--description extraction modules. This initial set provides the system with a foundation of plausible behaviours, but the search is not limited to them, where the archive can expand with newly discovered strategies, including tool calls proposed dynamically by LLMs.

Once initialised, the search process proceeds iteratively through a \emph{Design--Execute--Reflect} cycle (We provide the detailed procedure in Algorithm~\ref{alg:meddcr_summary}). At iteration $t$, a new workflow plan $\pi_t$ is proposed by an LLM-based Designer agent~\cite{li2023camel, wu2024autogen}, drawing on the memory archive that contains both the seeded workflows and prior designs $\mathcal{H}_{0:t-1}$. Each plan specifies a combination of tools, strategies, and reasoning steps, as well as their execution order. The plan is then compiled into an executable workflow $W_t$, translated into runnable code by a Coder agent, and subsequently executed to produce medical code predictions.

The execution results are assessed with an evaluation function $G(W_t)$, which integrates predictive performance (e.g., F1 score), guideline compliance~\cite{annicd}, and computational cost. Alongside the scalar score, the evaluation produces textual feedback produced by a \textit{Reflector} agent, diagnosing workflow errors such as ineffective tool combinations or guideline violations. Both the score and feedback are appended to the archive as a record $(\pi_t, W_t, s_t, r_t)$.

The memory archive thus grows with every iteration, providing two key signals for the search: (i) high-performing past workflows, which act as exemplars to imitate or refine, and (ii) diverse recent workflows, which encourage exploration. New proposals are therefore shaped both by the accumulated experience in the archive and by the evaluation feedback from prior iterations~\cite{hu2025automated}. Over time, the system learns to discover increasingly effective workflows by combining coding tools and strategies in novel ways.

The search loop terminates once the budget of iterations is reached or when performance converges. The final output is the best workflow $W^\star$ found in the archive, which can be used directly for automated medical coding or supplied as a strong starting point for further optimisation.

\subsection{Meta-Agent Architecture}
In this section, we present the details of the meta-agents within the MedDCR framework.
\paragraph{Designer Agent.} The Designer agent is responsible for generating candidate workflow plans $\pi$. At iteration $t$, it outputs
\begin{equation}
    \pi_t = \mathrm{Designer}(\mathcal{H}_t, \mathcal{L}, \Gamma) \in \Pi,
\end{equation}
where $\mathcal{H}_t$ is the memory archive, $\mathcal{L}$ is the component library (tools, strategies, and LLM modules), and $\Gamma$ encodes coding guidelines. The Designer operates under a structured \emph{meta-prompt} that (i) specifies constraints on how workflows must be formatted and executed, (ii) enumerates the available tool and strategy signatures, and (iii) appends informative exemplars from the memory archive.

In particular, the prompt shows the top-$k$ best-performing workflows and the most recent $n$ designs, ensuring that the Designer can exploit strong prior solutions while maintaining exploration. 

\paragraph{Coder Agent.} The Coder agent transforms abstract workflow plans into executable programs (e.g., Python-like codes). Given a plan $\pi_t$, it compiles the abstract plan into
\begin{equation}
    W_t = \mathrm{Coder}(\pi_t) \in \mathcal{S},
\end{equation}
where $W_t$ is an operational program that can be executed on clinical notes to generate medical codes.  

In practice, however, generated code may contain \emph{syntax or execution errors}, which would otherwise block evaluation. To address this, the Coder incorporates a \emph{self-fixing loop}: the executor first checks whether the compiled code runs successfully, if an error occurs, the Coder attempts to repair the code automatically and retries execution until a valid workflow is obtained or a retry budget is exceeded~\cite{joshi2023repair, zhang2024pydex}.  

This mechanism ensures that the search process is not derailed by syntactic inconsistencies~\cite{olausson2024is}. The separation of roles is thus preserved: the Designer explores high-level workflow planning, while the Coder guarantees that plans are translated into syntactically correct and executable implementations.

\paragraph{Reflector Agent.} The Reflector evaluates executed workflows and provides targeted feedback. Specifically, for each workflow $W_t$, it outputs
\begin{equation}
    (s_t, r_t) = \mathrm{Reflector}(W_t, D_{\text{val}}),
\end{equation}
where $s_t = G(W_t; D_{\text{val}}, \Gamma)$ is a scalar score that integrates predictive accuracy (e.g., F1 score or precision/recall), guideline compliance, and computational efficiency, and $r_t$ is a textual critique.  

To generate this feedback, the Reflector collects the intermediate outputs of each LLM call within the workflow and interprets them in the context of the corresponding operation. For example, it may highlight when an entity extraction step misses key mentions, when a guideline validator rejects a candidate code, or when reconciliation produces redundant outputs.  
The tuple $(\pi_t, W_t, s_t, r_t)$ is then appended to the archive, which is updated as
\begin{equation}
    \mathcal{H}_{t+1} = \mathcal{H}_t \cup \{(\pi_t, W_t, s_t, r_t)\}.
\end{equation}
This combination of quantitative scoring~\cite{khattab2024dspy} and fine-grained textual feedback~\cite{yang2023large, pryzant2023automatic} ensures that subsequent iterations improve not only raw performance but also the robustness and interpretability of workflows.

\begin{table*}[t]
\centering
\scriptsize
\setlength{\tabcolsep}{4pt}
\resizebox{0.9\textwidth}{!}{%
\begin{tabular}{c l ccc ccc}
\toprule
\multirow{2}{*}{\begin{tabular}[c]{@{}c@{}}Method Category \\(Label Space)\end{tabular}} & \multirow{2}{*}{Model} 
& \multicolumn{3}{c}{MDACE}
& \multicolumn{3}{c}{ACI-BENCH}\\
& & Precision & Recall & F1 & Precision & Recall & F1 \\
\midrule

\multirow{2}{*}{\textsc{\begin{tabular}[c]{@{}c@{}}PLM \\(1K)\end{tabular}}}
 & ICD~\cite{huang2022plm} &\textbf{0.49} &0.47  &\textcolor{gray}{\textbf{0.48}} &0.43  &0.41 & 0.42  \\
 & CA~\cite{douglas2025less} & \textcolor{gray}{\textbf{0.46}} &0.45  &0.45  &\textbf{0.44} &0.42 &0.43  \\

\arrayrulecolor{gray!40}\midrule
\multirow{3}{*}{\begin{tabular}[c]{@{}c@{}}Coder\\Workflow\\(1K)\end{tabular}}
 & RRS~\cite{kwan2024large} & 0.24 & 0.30 & 0.27 & 0.26 & 0.52 & 0.35  \\
 & MAC~\cite{li2024exploring} & 0.27 & 0.31 & 0.29 & 0.23  & 0.50  & 0.31 \\
 & CLH~\cite{motzfeldt2025code} & 0.45 & 0.40  & 0.42  & \textbf{0.44} & 0.39 & 0.41 \\

\arrayrulecolor{gray!40}\midrule
\multirow{6}{*}{\begin{tabular}[c]{@{}c@{}}Agentic\\Method\\ (1K)\end{tabular}}
 & CoT~\cite{wei2022chain} & 0.30 & 0.31 & 0.30 & 0.35 & 0.50 & 0.41 \\
 & CoT-SC~\cite{wangself} & 0.39 & 0.43 & 0.41 & 0.36 & 0.59 & 0.44 \\
 & MulDe~\cite{du2023improving} & 0.21 & 0.40 & 0.28 & 0.16 & 0.65 & 0.25 \\
 & Judge~\cite{zheng2023judging} & 0.26 & 0.53 & 0.35 & 0.22 & 0.64 & 0.33 \\
& MNER~\cite{Goel_LangExtract_2025} & 0.13 & 0.48 & 0.20 & 0.15 & \textcolor{gray}{\textbf{0.67}} & 0.25 \\
 & ADAS~\cite{hu2025automated} & 0.37 & 0.51 & 0.43 & 0.28 & 0.59 &0.43 \\
\arrayrulecolor{black}\midrule
\rowcolor{gray!8}
 & MedDCR-GPT-4o & 0.41 & \textcolor{gray}{\textbf{0.55}} & 0.47 & 0.36 & 0.65 & \textcolor{gray}{\textbf{0.46}}\\
\rowcolor{gray!8}
 & MedDCR-GPT-5 & \textcolor{gray}{\textbf{0.46}} & \textbf{0.59} & \textbf{0.51} & \textcolor{gray}{\textbf{0.43}} & \textbf{0.67} & \textbf{0.52}\\
\bottomrule
\end{tabular}}
\caption{\textbf{Main results on MDACE and ACI-BENCH datasets.}
The best results are highlighted in bold, and the second-best results are shown in \textcolor{gray}{\textbf{gray bold}}. Methods are grouped into three categories: Pretrained Language Models, expert-designed coding workflows, and agentic workflow strategies (including agent-based search methods).}
\label{tab:medical-coding-results}
\end{table*}

\subsection{Memory Archive and Plug-and-Play}
A central component of our proposed framework is the memory archive, which records the history of all explored workflows. At iteration $t$, the archive $\mathcal{H}_t$ contains tuples
\begin{equation}
\mathcal{H}_t = \{(\pi_j, W_j, s_j, r_j)\}_{j < t},
\end{equation}
where $\pi_j$ is the workflow plan proposed by the Designer, $W_j$ is the compiled executable workflow, $s_j$ is the score, and $r_j$ is the textual feedback.  

This archive plays a dual role: it enables \emph{reuse} of effective workflows by presenting the top-$k$ highest scoring exemplars to the Designer, and it supports \emph{exploration} by also presenting the $n$ most recent workflows, which prevent the search from collapsing into repetitive local optima~\cite{zhu2023ghost}. 

In this way, the Designer is encouraged to learn from successful past patterns while avoiding premature convergence to a single family of workflows~\cite{zhong2024memorybank}.  
As the search progresses, the archive grows into a structured memory of workflow designs, making it an adaptive optimiser that improves continuously over time.

Beyond storing internal proposals, the memory archive also supports a \emph{plug-and-play} mode. Here, external expert-crafted workflows, denoted $\mathcal{W}_{\text{seed}}$, are injected into the initial archive $\mathcal{H}_0$. These workflows may come from prior research, human-designed pipelines, or established clinical coding heuristics~\cite{gan-etal-2025-aligning,douglas2025less, motzfeldt2025code}.  
Once seeded, MedDCR treats them like any other entry in the archive: they can be directly reused, refined through feedback, or combined with newly generated strategies. 

This property makes our framework workflow-agnostic: it can either discover new workflows from scratch or improve upon existing designs, depending on the application scenario.  
The final output is therefore not limited to the best design discovered during search, but can also include optimised variants of expert workflows, providing a bridge between automated search and human expertise.

\begin{table*}[t]
\centering
\small
\renewcommand{\arraystretch}{1.2} 
\setlength{\tabcolsep}{6pt} 
\resizebox{\textwidth}{!}{%
\begin{tabular}{
    l 
    >{\centering\arraybackslash}p{2.5cm} 
    >{\centering\arraybackslash}p{2.5cm} 
    >{\centering\arraybackslash}p{2cm}   
    >{\centering\arraybackslash}p{2.5cm} 
    >{\centering\arraybackslash}p{2.5cm} 
    >{\centering\arraybackslash}p{2cm}   
}
\toprule
\multirow{2}{*}{Method} 
& \multicolumn{3}{c}{\textbf{MDACE}} 
& \multicolumn{3}{c}{\textbf{ACI-BENCH}} \\ 
\cmidrule(lr){2-4} \cmidrule(lr){5-7}
& Search Tokens
& Exec. Tokens
& Cost (USD)
& Search Tokens 
& Exec. Tokens
& Cost (USD) \\
\midrule
MedDCR
& 19{,}994 
& 11{,}545{,}586 
& \textbf{\$17.09} 
& 14{,}294 
& 2{,}233{,}623 
& \textbf{\$5.72} \\
\bottomrule
\end{tabular}}
\caption{\textbf{Computation Cost Comparison.} 
Token usage and projected cost in USD per 100 inference samples per search loop on MDACE and ACI-BENCH datasets. While effective, our method remains cost-efficient.}
\label{tab:cost_table}
\end{table*}
\section{Experiments}
\label{sec:experiments}
\begin{table}[t]
\centering
\small
\setlength{\tabcolsep}{6pt}
\begin{tabular}{lccc}
\toprule
\multirow{2}{*}{Seed Setting} & \multicolumn{3}{c}{MDACE} \\
\cmidrule(lr){2-4}
& Precision & Recall & Micro-F1 \\
\midrule
CoT-SC (seed only) & 0.37 & 0.48 & 0.42 \\
+ Quality-Diversity      & 0.38 & 0.48 & 0.42 \\
+ Multi-Debate     & 0.37 & 0.53 & 0.44 \\
+ NER          & 0.35 & 0.55 & 0.43 \\
+ All auxiliary    & \textbf{0.41} & \textbf{0.55} & \textbf{0.47} \\
\arrayrulecolor{gray!40}\midrule
No seed (scratch)  & 0.39 & 0.51 & 0.44 \\
\arrayrulecolor{black}\bottomrule
\end{tabular}
\caption{\textbf{Plug-and-play validation}: MedDCR initialised with CoT-SC as the primary seed, optionally augmented with auxiliary seeds (Self-Refine, Multi-Debate, Med-NER). Optimisation consistently improves the baseline CoT-SC workflow and outperforms search from scratch.}
\label{tab:plugplay}
\end{table}
\subsection{Experimental Setup}
\label{sec:exp_setup}
\paragraph{Dataset.} We evaluate our framework on two publicly available ICD-10 coding benchmarks. MDACE~\cite{cheng-etal-2023-mdace} provides expert-verified ICD-10 annotations to a mix of inpatient and professional-fee charts drawn from MIMIC-III~\cite{johnson2016mimic}, including discharge summaries, radiology reports and physician notes. ACI Benchmark~\cite{aci-bench} is a synthetic dataset of clinical notes paired with ICD-10 codes for benchmarking automated coding systems. Together, they enable evaluation of both coding accuracy and explainability in real-world settings.
\paragraph{Baseline.} We benchmark MedDCR against a broad set of existing approaches spanning \textit{three} paradigms. First, we consider pre-trained language model methods that frame coding as multi-label classification and require re-training on coding datasets, including PLM-ICD\cite{huang2022plm}, and PLM-CA~\cite{douglas2025less}. Second, we include expert-designed medical workflows that rely on heuristic rules or structured coding pipelines, represented by RRS~\cite{kwan2024large}, MAC~\cite{li2024exploring}, and CLH~\cite{motzfeldt2025code}. Finally, we evaluate against general agentic strategies and automated search frameworks that build on large language models with structured reasoning or multi-agent coordination, such as Chain-of-Thought~\cite{wei2022chain}, Self-Consistency~\cite{wang2023selfconsistency}, Multi-Debate~\cite{du2023improving}, Self-Refine~\cite{shinn2023reflexion}, and NER~\cite{Goel_LangExtract_2025}, as well as recent optimisation frameworks like ADAS~\cite{hu2025automated}. GPT-4o~\cite{hurst2024gpt} serves as the backbone model for all baseline agents. Together, these baselines span model-based, rule-based, and agentic paradigms, enabling a comprehensive comparison.
\paragraph{Implementations.} We use GPT-based models as the backbone for MedDCR, evaluating both GPT-4o~\cite{hurst2024gpt} and GPT-5~\cite{OpenAI}. The search loop is run for 100 iterations, with the Designer conditioned on the top-5 highest scoring and 3 most recent workflows from the archive at each step. To initialise the archive, we include all baselines from the aforementioned agentic workflow group, ensuring a diverse starting pool. Coding tools are modified and adapted from the simple-icd-10 library, covering both ICD-10-CM and PCS codes. More details are in the Appendix~\ref{app:meta}.
\paragraph{Evaluation Metrics.} We report standard measures for extreme multi-label coding: micro-F1, precision, and recall~\cite{huang2022plm, edin2024unsupervised, douglas2025less}. Micro-F1 balances overall precision and recall across the large code space, providing a primary measure of workflow effectiveness. Precision reflects the system’s ability to avoid spurious code assignments, while recall captures its capacity to recover the full set of relevant codes. Together, these metrics indicate how well the proposed workflows navigate the trade-off between exploration (capturing diverse correct codes) and exploitation (maintaining accuracy).

\begin{table}[t]
\centering
\small
\setlength{\tabcolsep}{6pt}
\begin{tabular}{lccc}
\toprule
& Precision & Recall & Micro-F1 \\
\midrule
MedDCR-GPT-4o        & 0.41 & 0.55 & 0.47 \\
\arrayrulecolor{gray!40}\midrule
-- No Top-$k$              & 0.35 & 0.42 & 0.38 \\
-- No Recent-$n$           & 0.33 & 0.34 & 0.34 \\
-- No Score Feedback       & 0.39 & 0.47 & 0.43 \\
-- No Text Feedback        & 0.44 & 0.50 & 0.47 \\
-- No Guideline in Meta    & 0.40 & 0.57 & 0.47\\
-- No Exemplar in Meta     & 0.36 & 0.51  &0.42 \\
\arrayrulecolor{black}\bottomrule
\end{tabular}
\caption{\textbf{Ablation study} on MDACE dataset. Each row removes one core component of MedDCR. The performance drops confirm the importance of the workflow exemplars, reflector feedback, guideline constraints and few-shot exemplar in achieving the full performance.}
\label{tab:ablation}
\end{table}

\begin{figure*}[t]
  \includegraphics[width=\textwidth, page=1]{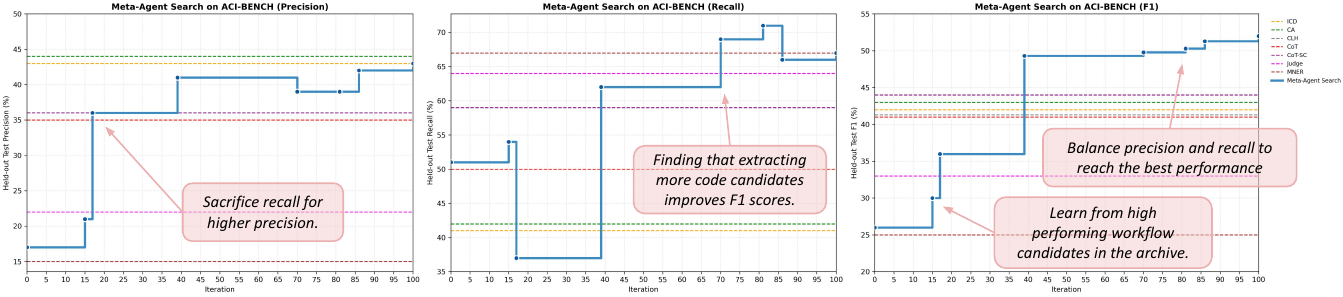}
  \caption{\textbf{Case study of the search process on ACI-Bench.} The blue line tracks the best workflow discovered at each iteration, measured by F1. The figure illustrates how performance improves as the system explores diverse candidates, learns from high-performing workflows, and balances precision and recall to refine the final design.}
  \label{fig:case}
\end{figure*}

\subsection{Main Results}
\label{sec:main_results}
As shown in Table~\ref{tab:medical-coding-results}, MedDCR consistently outperforms all baselines on both MDACE and ACI-BENCH. On MDACE, MedDCR-GPT-5 achieves a Micro-F1 of 0.51, improving over the strongest PLM baseline (ICD) and the best agentic baseline (CoT-SC) by 3\%. On ACI-BENCH, MedDCR-GPT-5 reaches 0.52 for F1, again surpassing PLM and agentic baselines. These results demonstrate that automated workflow search yields more effective coding strategies than either retrained language models or expert-crafted agentic pipelines. Importantly, Table~\ref{tab:cost_table} demonstrates that these improvements are achieved with modest overhead. Despite the iterative search process, MedDCR remains cost-efficient: a single search loop requires fewer than 20k search tokens on MDACE and 15k on ACI-BENCH, with total costs of \$17.09 and \$5.72 per 100 inference samples, respectively. Notably, the vast majority of cost arises from execution tokens rather than search tokens, showing that workflow optimisation overhead is modest relative to workflow execution.  Finally, Table~\ref{tab:plugplay} validates MedDCR’s plug-and-play property. Using CoT-SC as the primary seed, optimisation alone improves F1 from 0.41 to 0.42. When augmented with auxiliary seeds, performance increases further, reaching 0.47 F1 with all auxiliaries. Compared to search from scratch, seeded optimisation also achieves higher final scores. This confirms that MedDCR can flexibly refine and use existing workflows and diverse strategies, making it practical as a general workflow optimiser for medical coding.

\subsection{Ablation Studies}
\label{sec:ablation}
Table~\ref{tab:ablation} reports ablations on MDACE, where we systematically remove key components of MedDCR. Removing the top-$k$ or recent-$n$ workflows causes the largest performance drops of 0.09 and 0.13 in F1, confirming that both exploitation of strong designs and diversity from recent ones are essential for effective search. Feedback from the reflector is also critical: without score feedback, F1 falls to 0.43, while removing textual feedback prevents further improvements despite retaining numeric scores. Guideline constraints show a trade-off, slightly increasing recall but offering no net F1 gain when excluded. Finally, removing all exemplars from the meta-prompt reduces F1 to 0.42, underscoring their role in stabilizing the workflow generation. Overall, the results highlight that MedDCR’s gains arise from the complementary contributions of memory selection, reflective feedback, guideline grounding, and exemplar conditioning.
\subsection{Case Study}
\label{sec:case_study}
Figure~\ref{fig:case} illustrates how our framework improves over the course of the search in terms of precision, recall, and F1, with the blue line indicating the best workflow discovered at each iteration as measured by F1. In the early stages, the search explores relatively simple strategies, while later iterations integrate more complex tool use and validation steps, resulting in steady performance gains. This progression highlights the role of memory and reflective feedback in escaping suboptimal designs and converging toward stronger workflows. The final best workflow combines multiple reasoning strategies and coding tools in a coordinated sequence, exemplifying the kind of designs that MedDCR can automatically uncover. Due to space, the best workflow structure is provided in the Appendix~\ref{app:case}.
\section{Conclusions}
\label{sec:conclusion}
We presented MedDCR, an automated framework for workflow optimisation in medical coding. Unlike prior work that fixes workflow structures by expert design, MedDCR treats workflow construction as a learning problem, realised through iterative design–execute–reflect loops. Central to the framework is a meta-agent architecture with a Designer, Coder, and Reflector, supported by a memory archive that enables both search from scratch and plug-and-play refinement of expert workflows. Experiments on MDACE and ACI-BENCH demonstrate that MedDCR achieves consistent improvements over state-of-the-art baselines, while also providing interpretable feedback through the Reflector. These results show that automated workflow optimisation can yield both higher accuracy and greater reliability than static, manually designed systems, offering a principled approach to building trustworthy medical coding agents.
\section{Limitations}
While MedDCR demonstrates strong performance and adaptability, several limitations remain. First, the framework relies on general-purpose GPT models~\cite{hurst2024gpt,OpenAI} that are not fine-tuned for medical coding, which may limit their understanding of domain-specific terminology and guideline nuances. A medically fine-tuned or instruction-adapted foundation model could further improve both accuracy and compliance with coding rules. Second, workflow optimisation is computationally expensive~\cite{hu2025automated}, as it involves repeated LLM executions and feedback loops~\cite{agrawal2025gepa}. Although the search cost is moderate relative to inference, scaling to larger datasets or real-time settings may require more efficient search strategies~\cite{calvaresi2021real}. Third, the textual feedback generated by the Reflector is heuristic and may occasionally produce ambiguous or overly general guidance, suggesting a need for structured reflection or rule-grounded critics~\cite{dong2025rag}. Fourth, the current framework operates purely on textual information, incorporating multimodal signals such as visual~\cite{willemink2020preparing, zheng2022towards, tu2024ranked} or structured clinical data~\cite{tayefi2021challenges},could enhance its ability to capture richer diagnostic cues and improve both predictive accuracy and cross-domain generalisation. Finally, our experiments focus primarily on ICD-10 coding, extending MedDCR to multi-lingual settings~\cite{miranda2020codiesp} would provide stronger evidence of generality. Addressing these limitations would bring MedDCR closer to deployment ready, specialised agentic systems for clinical applications.

\bibliography{custom}

\appendix

\section*{Appendix}
\label{sec:appendix}
\section{Medical Coding Background}
\noindent
\textbf{Medical coding task.}  
Medical coding is the process of translating unstructured clinical documentation into standardized diagnostic and procedural codes, most commonly following the International Classification of Diseases (ICD) standard. Each clinical note may contain multiple diagnoses, procedures, and contextual information that must be mapped to corresponding ICD-10-CM (diagnosis) and ICD-10-PCS (procedure) codes. The task is inherently \textit{multi-label} and \textit{context-dependent}.

\vspace{0.5em}
\noindent
\textbf{Example.}  
Consider the following excerpt from a fabricated discharge summary:
\begin{quote}
\textit{“The patient was admitted for acute myocardial infarction and underwent coronary artery bypass graft (CABG). The postoperative course was uncomplicated.”}
\end{quote}
A correct ICD-10-CM coding outcome for this note includes:
\begin{itemize}
    \item \texttt{I21.3} — ST elevation (STEMI) of unspecified site.
\end{itemize}
and for ICD-10-PCS (procedure codes):
\begin{itemize}
    \item \texttt{021009W} — Bypass coronary artery, one site, autologous vein, open approach.
\end{itemize}

\vspace{0.5em}
\noindent
\textbf{Challenges.}  
Accurate code assignment requires multiple reasoning steps:
\begin{enumerate}
    \item Identifying relevant clinical entities (e.g., diseases, procedures).
    \item Consulting external resources such as the ICD alphabetic index and tabular list.
    \item Applying coding guidelines (e.g., sequencing, laterality, and exclusion rules).
    \item Cross-checking consistency between diagnoses and procedures.
\end{enumerate}
For example, a note describing “Type 2 diabetes with chronic kidney disease” must yield both the primary diagnosis (\texttt{E11.22}) and the complication (\texttt{N18.9}) while respecting guideline-specific linkage rules.

\vspace{0.5em}
\noindent
\textbf{Motivation for automation.}  
Given the need for cross-referencing multiple resources and enforcing rule-based logic, medical coding naturally lends itself to \textit{workflow-based} reasoning rather than direct text classification. The MedDCR framework aims to automate this multi-step reasoning process by discovering workflows that can effectively combine tool use, rule application, and reflection.

\section{Data Consent and Usage}

\noindent
All datasets used in this study are derived from publicly available, de-identified clinical corpora with appropriate data use agreements. The MDACE dataset~\cite{cheng-etal-2023-mdace} is constructed from the MIMIC-III database~\cite{johnson2016mimic}, which contains de-identified electronic health records from the Beth Israel Deaconess Medical Center. Access to MIMIC-III requires completion of the PhysioNet credentialing process and acceptance of a data use agreement that prohibits any attempt at patient re-identification. Our use of MDACE fully complies with these requirements.  

The ACI-BENCH dataset~\cite{aci-bench} is publicly released under an open-access license and contains synthetic or anonymized clinical notes for benchmarking purposes. No protected health information (PHI) was accessed or processed in any part of this study. All experiments were conducted in accordance with institutional data governance and ethical use policies.

\begin{algorithm}[!htbp]
\caption{Evidence-Aware Coding Pipeline}
\label{alg:pipeline}
\begin{algorithmic}[1]
\Require Clinical note $N$; hyperparameters: $T_{\text{terms}}$ (term extractor), $S{=}4$, $S_{\max}{=}2$, $W{=}24$, $\tau_{\text{keep}}{=}0.5$, $\gamma{=}0.15$, $K{=}20$
\Ensure Ranked list $\mathcal{R}$ of codes with scores and evidence

\State $E \gets \textsc{Terms}(\textsc{ReasonForVisit}(N); T_{\text{terms}})$ \Comment{extract terms/entities w/ reason-for-visit cues}
\State $C_{s} \gets \textsc{SearchAlphaIndex}(\textsc{SynExp}(E))$
\State $C_{t} \gets \textsc{ProposeFromTerms}(E; \text{mode}=\text{plain})$
\State $C_{tc} \gets \textsc{ProposeFromTermsCoT}(E; \text{ref}= \text{sec} ; \text{mode}=\text{CoT})$
\State $C_{n} \gets \textsc{ProposeFromNote}(\textsc{RFV}(N); \text{mode}=\text{plain}; \text{samples}=S)$
\State $C_{nc} \gets \textsc{ProposeFromSec}(N; \text{mode}=\text{CoT}, \text{samples}=S)$
\State $\hat{C} \gets \textsc{Merge}(C_{s}, C_{t}, C_{n}, C_{tc}, C_{nc})$ \Comment{set union with deduplication}
\State $\hat{C} \gets \textsc{Canonicalize}(\hat{C})$
\State $\hat{C} \gets \textsc{Validate}(\hat{C})$ \Comment{schema/rule compliance; remove invalid codes}
\State $D \gets \textsc{FetchDesc}(\hat{C})$ \Comment{$D:$ map code $\mapsto$ description from tabular index}
\State $m_{\text{desc}} \gets \textsc{DescMatch}(\hat{C}, D, N)$ \Comment{per-code description$\leftrightarrow$note match score}

\State $Z \gets \textsc{EvidenceLink}(N, \hat{C}; S_{\max}, W)$ 
\Comment{$Z:$ map code $\mapsto$ at most $S_{\max}$ snippets (window $W$ tokens)}

\State $S_{\text{judge}} \gets \textsc{Judge}\!(\hat{C}, Z; \ 
\text{judge\_strategy}=\text{per-code evidence-aware keep/drop}, \  
\text{output}=\text{json\_scores})$
\Comment{score in $[0,1]$ for each code}

\State $\tilde{C} \gets \{\, c \in \hat{C}\ :\ S_{\text{judge}}(c) \ge \tau_{\text{keep}} \,\}$ 
\Comment{filter by judge score}

\State $\tilde{C} \gets \textsc{ContrastiveScreen}\!(\tilde{C};\ \text{by}=\text{sibling\_desc},\ \text{contrast\_margin}=\gamma, \text{action}=\text{drop\_or\_demote})$\
\Comment{prune/demote near-duplicate siblings with small description margin}

\ForAll{$c \in \tilde{C}$}
\vspace{-1em}
  \begin{align*}
  &s(c) \gets \textsc{RerankScore}(S_{\text{judge}}(c), \\
   &m_{\text{desc}}(c), \textsc{EvidenceStrength}(Z(c)))
  \end{align*}
\vspace{-2em}
\EndFor
\State $\mathcal{R} \gets \textsc{SortBy}(\tilde{C},\ \text{key}=s,\ \text{desc})$

\If{$|\mathcal{R}| < K$}
  \State $\mathcal{B} \gets \textsc{FallbackTopK}(\hat{C}\setminus \tilde{C},\ K-|\mathcal{R}|)$
  \State $\mathcal{R} \gets \textsc{Concat}(\mathcal{R},\ \mathcal{B})$
\EndIf

\State \Return $\textsc{ReturnRaw}(\mathcal{R},\ s,\ Z,\ m_{\text{desc}})$
\end{algorithmic}
\end{algorithm}

\section{Case Study and Pseudo-Code of the Searched Workflow} \label{app:case}

\noindent
\textbf{Overview.}  
To better illustrate how MedDCR operates in practice, we present the best workflows discovered on the ACI-Benchmark dataset. This case study corresponds to the pointed workflow in Figure~\ref{fig:case}, showing how iterative search progressively refines design choices through reflective feedback and archive-guided learning (Algorithm~\ref{alg:pipeline} and ~\ref{alg:pipeline-2}).

\vspace{0.5em}
\noindent
\textbf{Workflow evolution.}  
In early iterations, the Designer generated linear pipelines focused mainly on entity extraction and description matching. As search progressed, reflective feedback encouraged the integration of verification and reconciliation steps, such as cross-checking alphabetic and tabular index results or re-validating low-confidence predictions. The final workflow incorporates diagnostic reasoning, combining multiple coding tools with validation and guideline enforcement to ensure consistency for the final predictions.

\vspace{0.5em}
\noindent
\textbf{Pseudo-code of the searched workflow.}
The pipeline balances recall and precision through a two-stage process. In the first stage (Lines 1–7, Algo~\ref{alg:pipeline}), it expands recall by generating a large pool of candidate codes from multiple sources including alphabetic index lookup with extracted terms (expanded with synonyms), term-based, and note-based proposals from LLMs, ensuring broad coverage. In the second stage, precision is progressively enforced through validation, evidence linking, and filtering. Invalid codes are removed via rule checks, an evidence-aware judge filters low-confidence candidates based on $\tau_{keep}$, and contrastive screening $\gamma$ prunes near-duplicates. Finally, reranking integrates judge scores, description similarity, and evidence strength to prioritize high-precision codes while maintaining recall from the initial expansion.

\begin{algorithm}[H]
\caption{Evidence- and Vote-Aware Coding}
\label{alg:pipeline-2}
\begin{algorithmic}[1]
\Require Clinical note $N$; $S_t{=}3$,  $S_n{=}4$, $S_{\max}{=}2$, $M{=}256$, $\theta{=}0.5$, $K{=}20$
\Ensure Ranked list $\mathcal{R}$ of codes with scores and evidence

\State $E \gets \textsc{Terms}(N)$
\State $C_{t}^{(1:S_t)} \gets \textsc{ProposeFromTerms} \newline (E;\ \text{mode}=\text{plain},\ \text{samples}=S_t)$
\State $C_{n}^{(1:S_n)} \gets \textsc{ProposeFromNote} \newline (N;\ \text{mode}=\text{CoT},\ \text{samples}=S_n)$
\State $\hat{C} \gets \textsc{Merge}\!\Big(\bigcup_{s=1}^{S_t} C_{t}^{(s)},\ \bigcup_{s=1}^{S_n} C_{n}^{(s)}\Big)$
\State $\hat{C} \gets \textsc{Canonicalize}(\hat{C})$
\State $\hat{C} \gets \textsc{Validate}(\hat{C})$\Comment{Self-consistency via normalized vote frequency}
\State $\text{votes}(c) \gets \newline \sum_{s=1}^{S_t} \mathbf{1}\{c\in C_{t}^{(s)}\} + \sum_{s=1}^{S_n} \mathbf{1}\{c\in C_{n}^{(s)}\}$ for $c\in\hat{C}$
\State $\text{vote\_ratio}(c) \gets \dfrac{\text{votes}(c)}{S_t + S_n}$; \quad annotate as ``vote\_ratio''

\State $D \gets \textsc{FetchDesc}(\hat{C})$ \Comment{$D$: code $\mapsto$ tabular description}
\State $m_{\text{desc}} \gets \textsc{DescMatch}(\hat{C}, D, N)$ \Comment{per-code description$\leftrightarrow$note match score in $[0,1]$}
\State $Z \gets \textsc{EvidenceExtract}\!(N, \hat{C};\ \text{strategy}=\text{snippet\_from\_note},\ \text{per\_code\_spans}=S_{\max},\ \text{max\_tokens}=M)$
\State $\text{evidence\_overlap}(c) \gets \textsc{EvidenceOverlap}(Z(c), N)$; \quad annotate as ``evidence\_overlap''
\State $\left(\text{judge\_keep}(c),\ \text{judge\_conf}(c)\right) \gets \textsc{Judge}\!(c, Z(c), D(c);\ \text{strategy}=\text{evidence\_tabular\_desc})$ for $c\in\hat{C}$
\State annotate $\text{judge\_conf}$ as ``judge\_conf''

\State $\tilde{C} \gets \{\, c \in \hat{C}\ :\ \text{judge\_keep}(c)\ \lor\ \text{judge\_conf}(c)\ge \theta \,\}$

\State $\tilde{C} \gets \textsc{HierPrune}\!(\tilde{C};\ \text{rules}=\{\text{prefer\_specific\_over\_unspecified},\newline \text{drop\_duplicate\_laterality},\newline \text{drop\_mutually\_exclusive\_with\_lower\_conf}\})$
\State \ForAll{$c \in \tilde{C}$}
  \State $s(c) \gets 0.4\cdot \text{vote\_ratio}(c) + 0.3\cdot m_{\text{desc}}(c) + 0.2\cdot \text{judge\_conf}(c) + 0.1\cdot \text{evidence\_overlap}(c)$
\EndFor
\State $\mathcal{R} \gets \textsc{SortBy}(\tilde{C},\ \text{key}=s,\ \text{desc})$

\If{$|\mathcal{R}| < K$}
  \State $\mathcal{B} \gets \textsc{FallbackTopK}(\hat{C}\setminus \tilde{C},\ K-|\mathcal{R}|)$
  \State $\mathcal{R} \gets \textsc{Concat}(\mathcal{R},\ \mathcal{B})$
\EndIf

\State \Return $\textsc{ReturnRaw}(\mathcal{R},\ s,\ Z,\ \text{vote\_ratio},\newline m_{\text{desc}}, \ \text{judge\_conf},\ \text{evidence\_overlap})$
\end{algorithmic}
\end{algorithm}

\section{Meta-Prompt and Coding Tools} \label{app:meta}

\subsection*{Meta-Prompt for the Designer Agent}

\noindent
The Designer agent is instructed through a meta-prompt that defines its role, input exemplars, design constraints, and expected JSON output format. The prompt dynamically incorporates the top-$k$ best-performing and $n$ most-recent workflows from the memory archive. Below is the formatted version of the prompt used in this study.

\begin{minipage}{\linewidth}
\begin{lstlisting}[style=prompt, language={}, caption={Meta-prompt of Designer Agent}]
You are designing a NEW controller for ICD-10 multi-label medical coding.
STAGE A (THIS TURN): PROPOSE A PLAN ONLY - DO NOT WRITE CODE.
Return a compact JSON plan that the system will implement later.

Think creatively while maintaining coding accuracy.

Rules:
- The plan is an ordered list of steps; each step is an object with an operation key ("op")
  and its required parameters.
- Exclude codes related to family history, negative, hypothetical, or ruled-out mentions.
- Include key parameters when relevant (e.g., samples, threshold, k, strategy, by).
- Avoid repeating recent workflow families unless changes are expected to improve F1.
- Keep the workflow concise and executable; the system will validate it.
- You may combine or extend existing operations if logically beneficial.

Available Tool Signatures:
MedicalTermExtraction(note) - Extract medical terms from the medical notes using LangExtract.
TabularIndexSearch(entity)   - Retrive code description from tabular index.
DescMatch(codes, note)       - Compute similarity between code descriptions and note text.
GuidelineValidator(codes, ruleset) - Filter codes violating ICD-10 rules.
Reconciler(codes)                 - Merge duplicates or conflicting predictions.
EvidenceLinker(note, codes, window, max_snips) - Extract evidence snippets per code.
Judge(codes, evidence, strategy)  - Assign per-code confidence scores.
...

Exemplars (for diversity & performance):
First are top performers by F1 (TOP_i), followed by the most recent attempts (RECENT_i).
Use these to inspire but not copy - propose an improved or novel workflow.

Top_1:
[Exemplar Workflow i]
...

Recent_1:
[Exemplar Workflow j]
...

Deliverable (Stage A): Return STRICTLY ONE JSON OBJECT (no prose):
{
  "name": "<controller name>",
  "thought": "<why this plan should improve F1>",
  "plan": [ { "op": "...", "...": "..." }, ... ]
}
If you introduce new operations, name them clearly - the system will map or extend the op set.
DO NOT include executable code in this turn.
\end{lstlisting}
\end{minipage}

\noindent
This meta-prompt encourages creativity while constraining the Designer to produce executable, guideline-compliant workflow plans. Tool signatures are embedded to ensure awareness of available operations and their expected parameters.

\subsection*{Meta-Prompt for the Coder Agent}

\noindent
The Coder agent translates workflow plans from the Designer into executable Python code that implements the proposed controller. The prompt strictly enforces the plan order, ensures syntactic validity, and outputs JSON-formatted code only. The full formatted prompt is shown below.

\begin{minipage}{\linewidth}
\begin{lstlisting}[style=prompt, language={}, caption={Meta-prompt of Coder Agent}]
Stage B: Implement code that EXACTLY follows the accepted plan.

Implement `class Controller` for ICD-10 multi-label coding by following this accepted plan JSON:
<INSERT PLAN JSON HERE>

Constraints:
 - Keep EXACT signature:
       class Controller:
           def forward(self, task)
 - Honour the plan order and parameters (thresholds, temperatures, samples, k, strategies, etc.).
 - Do NOT include any comments in code.
 - Return STRICT JSON ONLY (no prose, no markdown):
   {
     "code": "class Controller:\n    def forward(self, task):\n        ..."
   }

Implementation guidance:
 - For LLM calls, use LLMAgentBase.run_text with explicit roles and keep temperature/loops consistent with the plan.
 - Always de-duplicate outputs.

 === REFERENCE EXAMPLES (for guidance only; DO NOT change your plan) ===
 Reference plan JSON:
   <example plan JSON>

 Reference full implementation (follows the reference plan):
   <example implementation code>

 Additional example snippets:
   <short code fragments for guidance>
\end{lstlisting}
\end{minipage}

\noindent
The Coder’s self-fixing loop detects non-executable code via Python traceback parsing and regenerates corrected versions until a valid workflow is obtained.

\subsection*{Meta-Prompt for the Reflector Agent}
\noindent
The Reflector agent evaluates workflow executions, producing both quantitative scores and textual feedback. It receives the workflow plan, predictions, and gold codes, and is responsible for scoring and analysing workflow effectiveness. The structured prompt used is shown below.

\begin{minipage}{\linewidth}
\begin{lstlisting}[style=prompt, language={}, caption={Meta-prompt of Reflector Agent}]
You are the Reflector agent in the MedDCR framework.

Your goal is to evaluate the performance of a medical coding workflow
and provide concise feedback that helps the Designer improve in the next iteration.

Inputs:
 - Workflow plan JSON describing the sequence of operations.
 - Model predictions with supporting evidence spans.
 - Ground-truth ICD-10 codes from the validation set.
 - Quantitative scores (precision, recall, F1) computed for this workflow.

Tasks:
 1. Verify the provided scores for correctness and internal consistency.
 2. Identify the workflow's strengths and weaknesses in terms of tool use,
    reasoning order, and guideline adherence.
 3. Analyse failure cases such as:
      * incorrect entity extraction,
      * guideline violations,
      * missing or redundant validation steps,
      * low-confidence or conflicting predictions.
 4. Provide concise textual feedback that diagnoses the cause of errors
    and suggests actionable refinements (e.g., add validation, change order of tools, adjust threshold).
 5. Maintain objectivity-do not modify the workflow plan or fabricate new scores.

Return STRICT JSON ONLY (no prose, no markdown):
{
  "score": {
      "precision": <float>,
      "recall": <float>,
      "f1": <float>
  },
  "feedback": <str>"
}

Example feedback:
<Example Feedback>
\end{lstlisting}
\end{minipage}

\noindent
The textual feedback is appended to the archive and shown to the Designer in the next iteration, enabling reflective learning.

\subsection*{ICD-10 Coding Tool List}

\noindent
The framework provides a curated library of coding tools accessible to both the Designer and Coder agents. These tools are mainly adopted from the simple-icd-10-cm resource for CM codes\footnote{A full list of tools is available at \url{https://github.com/StefanoTrv/simple_icd_10_CM}}, e.g.,
\begin{itemize}
    \item \texttt{get\_parent} - returns the immediate parent of a code in the ICD-10-CM hierarchy; \texttt{prioritize\_blocks} disambiguates codes that can denote either a block or a category. 
    \item \texttt{get\_children} - returns the immediate children of a code; if a code name is ambiguous (both a block and a category), setting \texttt{prioritize\_blocks=True} treats it as the block (e.g., “B99” example). 
    \item \texttt{get\_ancestors} - returns all ancestor nodes (e.g., category to block to chapter) of the given code, with optional block prioritization to resolve ambiguity. 
    \item \texttt{get\_descendants} - returns all descendant nodes of the given code, with optional block prioritization for ambiguous names. 
    \item \texttt{is\_ancestor} - returns \texttt{True} iff \texttt{a} is an ancestor of \texttt{b}; optional flags control how to interpret ambiguous block/category codes. 
    \item  \texttt{is\_descendant} - returns \texttt{True} iff \texttt{a} is a descendant of \texttt{b}; with the same ambiguity controls. 
    \item \texttt{get\_nearest\_common\_ancestor} - returns the nearest common ancestor of \texttt{a} and \texttt{b} (or empty string if none), with optional block/category disambiguation. 
    \item \texttt{is\_leaf} - returns \texttt{True} iff the code is a leaf (has no children) in the ICD-10-CM hierarchy; supports block prioritization.
    \item \texttt{code\_to\_text} - returns the textual description associated with a given code.
    \item \texttt{text\_to\_codes} - maps natural-language text to candidate ICD-10-CM codes (by matching code descriptions).
    \item \texttt{load\_taxonomy} - loads the internal ICD-10-CM taxonomy data structure for lookups and hierarchies.
    \item \texttt{is\_valid\_code} - checks whether the given ICD-10-CM code exists in the taxonomy.
\end{itemize}
and we additionally create more complex LLM-based tools for both ICD-10-CM (diagnoses) and ICD-10-PCS (procedures):

\begin{itemize}
    \item \texttt{MedicalTermExtraction} - retrieves medical terms from the note using Langextract.
    \item \texttt{TabularIndexSearch} - retrieves hierarchical and related codes from the tabular list.
    \item \texttt{DescMatch} - computes similarity between ICD descriptions and note text.
    \item \texttt{GuidelineValidator} - applies ICD-10 coding rules to remove invalid or conflicting codes.
    \item \texttt{Reconciler} - merges duplicates and resolves inter-tool inconsistencies.
    \item \texttt{EvidenceLinker} - extracts supporting text snippets for each predicted code.
    \item \texttt{Judge} - assigns per-code confidence scores and keep/drop decisions.
\end{itemize}

Beyond the predefined toolset, the Designer agent is explicitly encouraged to invent new tools using the same LLM interface (\texttt{LLMAgentBase.run\_text}). When proposing new operations, the Designer must specify their role, purpose, and expected input/output signatures. This design encourages creative exploration of novel reasoning and validation strategies, allowing MedDCR to continuously expand its operational toolkit beyond the initial ICD-10 functions.

\begin{itemize}
    \item \texttt{LLMAgentBase.run\_text(prompt, role, loops)} - performs controlled text generation for tasks such as code proposal, evidence linking, or reflection. Explicit roles (e.g., ``coder'', ``judge'', ``reflector'') are specified, and parameters such as \texttt{loops} are kept consistent with the workflow plan.
\end{itemize}

\noindent
These tools collectively enable workflow designs that emulate professional coding processes, combining retrieval, validation, and reasoning for robust code prediction.

\section{Potential Risks}

\noindent
While MedDCR is designed to improve the reliability of automated medical coding, several potential risks should be acknowledged. First, as the underlying GPT-based models are not clinically validated or fine-tuned for medical reasoning, they may generate inaccurate or incomplete workflows that lead to incorrect code assignments. Such errors could propagate in downstream analyses or billing if deployed without human oversight. Second, reflective feedback generated by large models may include unverifiable or inconsistent rationales, which, if misinterpreted, could undermine trust in automated explanations. Third, because workflow optimisation relies on data derived from de-identified clinical corpora, there remains a residual risk of bias reflecting documentation practices from specific institutions or regions. Finally, automated systems like MedDCR should not be used for direct clinical decision-making or patient care without rigorous validation and professional review. These risks highlight the importance of using MedDCR strictly for research and development under appropriate governance and human-in-the-loop supervision.

\end{document}